\documentclass[letterpaper, 10 pt, journal, twoside]{IEEEtran}

\ifCLASSINFOpdf
\else
\fi

\usepackage{balance}
\usepackage{color}

\usepackage{graphics} 
\usepackage{amsmath,amssymb,amsfonts} 
\usepackage{bm}
\usepackage{subcaption}	 
\usepackage{graphicx}
\usepackage{multirow}
\usepackage{array}

\usepackage{url}
\usepackage{hyperref}
\usepackage{xcolor}

\usepackage{booktabs}
\usepackage{multirow}
\usepackage{textcomp}
\usepackage{setspace}
\usepackage{tabularx}
\usepackage{makecell}
\usepackage{colortbl}  
\usepackage{pifont}
\newcommand{\cmark}{\ding{51}}
\newcommand{\xmark}{\ding{55}}

\newcommand{\methodname}{LiCROcc}
\newcommand{\ie}{\textit{i.e.}}
\definecolor{barrier}{RGB}{112,128,144}
\definecolor{bicycle}{RGB}{220,20,60}
\definecolor{bus}{RGB}{255, 127, 80}
\definecolor{car}{RGB}{255, 158, 0}
\definecolor{const. veh.}{RGB}{233, 150, 70}
\definecolor{motorcycle}{RGB}{255,61,99}
\definecolor{pedestrian}{RGB}{0,0,230}
\definecolor{traffic cone}{RGB}{47,79,79}
\definecolor{trailer}{RGB}{255,140,0}
\definecolor{truck}{RGB}{255,99,71}
\definecolor{drive. suf.}{RGB}{0,207,191}
\definecolor{other flat}{RGB}{175,0,75}
\definecolor{sidewalk}{RGB}{75,0,75}
\definecolor{terrain}{RGB}{112,180,60}
\definecolor{manmade}{RGB}{222,184,135}
\definecolor{vegetation}{RGB}{0,175,0}

\usepackage[backend=bibtex,natbib=true,style=numeric-comp,sorting=none,giveninits=true,maxbibnames=99,url=false,doi=false]{biblatex}
\begin{document}
\title{LiCROcc: Teach Radar for Accurate Semantic Occupancy Prediction using LiDAR and Camera}

\author{Yukai Ma$^{1,2, \dagger}$, Jianbiao Mei$^{1,2, \dagger}$, Xuemeng Yang$^{2}$, Licheng Wen$^{2}$, Weihua Xu$^{1}$, \\Jiangning Zhang$^{1}$, Botian Shi$^{2,*}$, Yong Liu$^{1,*}$, Xingxing Zuo$^{3}$ 
\thanks{$^{1}$The authors are with the Institute of Cyber-Systems and Control, Zhejiang University, Hangzhou, China.}%
\thanks{$^{2}$The authors are with the Shanghai Artificial Intelligence Laboratory, Shanghai, China.}%
\thanks{$^{3}$The author is with the Department of Computer Engineering, Technical University of Munich, Germany}%
\thanks{$^{*}$Botian Shi and Yong Liu are the corresponding authors (Email: shibotian@pjlab.org.cn; yongliu@iipc.zju.edu.cn)}%
\thanks{$^\dag$The co-first authors have equal contributions.}
}
\maketitle

\begin{abstract}
Semantic Scene Completion (SSC) is pivotal in autonomous driving perception, frequently confronted with the complexities of weather and illumination changes. The long-term strategy involves fusing multi-modal information to bolster the system's robustness. Radar, increasingly utilized for 3D target detection, is gradually replacing LiDAR in autonomous driving applications, offering a robust sensing alternative.
In this paper, we focus on the potential of 3D radar in semantic scene completion, pioneering cross-modal refinement techniques for improved robustness against weather and illumination changes, and enhancing SSC performance.Regarding model architecture, we propose a three-stage tight fusion approach on BEV to realize a fusion framework for point clouds and images. 
Based on this foundation, we designed three cross-modal distillation modules—CMRD, BRD, and PDD. 
Our approach enhances the performance in both radar-only (R-LiCROcc) and radar-camera (RC-LiCROcc) settings by distilling to them the rich semantic and structural information of the fused features of LiDAR and camera.
Finally, our LC-Fusion (teacher model), R-LiCROcc and RC-LiCROcc achieve the best performance on the nuScenes-Occupancy dataset, with mIOU exceeding the baseline by 22.9\%, 44.1\%, and 15.5\%, respectively. The project page is available at \url{https://hr-zju.github.io/LiCROcc/}.
\end{abstract}


%
\IEEEpeerreviewmaketitle

%
%
%
%

\section{INTRODUCTION}
Semantic Scene Completion (SSC), a crucial technology in autonomous driving, has garnered substantial attention for its ability to ground detailed 3D scene information. Cameras and LiDAR are the most prevalent sensors used for SSC tasks, each with strengths and limitations. The former provides rich semantic context but lacks depth information and is susceptible to lighting and weather conditions. The latter offers accurate 3D geometry but performs poorly when given highly sparse input, and is hindered for wide applications due to the high cost of dense LiDAR sensors.
On the other hand, radar, a weather-resistant sensor gaining traction in autonomous driving, is valued for its automotive-grade design and affordability. Despite its robustness in diverse weather and lighting conditions, radar's sparse and noisy measurements present significant challenges for SSC in large-scale outdoor scenarios. 

\begin{figure}[t]
    \captionsetup{font={small}}
    \centering
    \includegraphics[width=1 \linewidth]{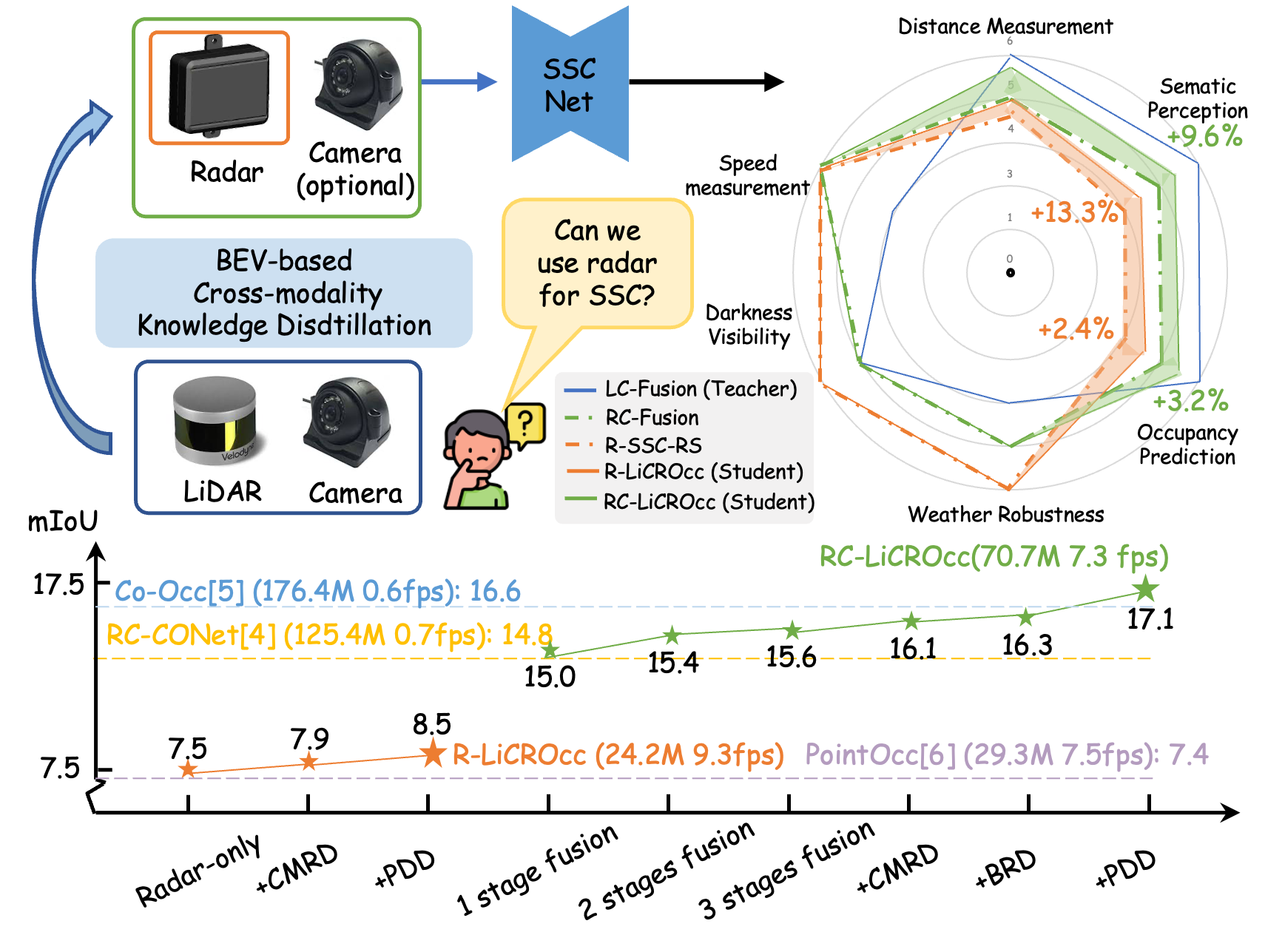}
    \caption{Our motivation for proposing LiCROcc. We explore radar performance on SSC tasks with the expectation of developing a network with balanced performance and robustness. We further improve radar sensors' semantic occupancy prediction and distance measurement capabilities by cross-modal distillation while maintaining their inherent night vision capabilities and weather robustness. In the figure on the right, the blue line represents the capability of the teacher model, the dashed line represents the capability of the pre-distillation student in our two settings, and the solid line in the same color represents the equilibrium performance of the model after KD.
    The graphs on the bottom demonstrate the enhancement of our method after gradually adding modules and the comparison with other methods on nuScenes-Occupancy~\cite{wang2023openoccupancy} (validation set), including Co-Occ~\cite{pan2024co}, RC-CoNet~\cite{wang2023openoccupancy}, and PointOcc~\cite{zuo2023pointocc}.
    }
    \label{fig:wallpaper}
\vspace{-10pt}
\end{figure}

Most of the research has recently focused mainly on radar-based detection~\cite{kim2023crn,long2023radiant}. Only a few studies~\cite{ding2024radarocc} have explored the application of radar sensors in the SSC task. However, they can only use radar to predict occupancy in very few categories or as a supplement to multi-modal inputs. In addition, we found that although the radar has inherent strengths against adverse weather conditions and illumination changes, as indicated in Tab.~\ref{table:base_main} and Fig.~\ref{fig:wallpaper}, there is still a significant performance gap between the radar-based and LiDAR/camera-based SSC approaches.
To address the above challenge, in this work, we explore using radar as a core sensor for SSC and set a new sota regarding the performance.

Initially, we establish a radar-based baseline R-SSC-RS inspired by~\cite{mei2023ssc}. However, due to the lack of intricate details, relying solely on radar sensors cannot achieve highly accurate and robust SSC. Therefore, to improve the performance of the radar-based SSC, we further design a camera-radar fusion network to fuse the rich semantic context of RGB images into radar in the BEV space effectively and efficiently. In this way, the radar SSC has been significantly improved.

Moreover, we have observed that LiDAR-camera fusion achieves superior performance in outdoor SSC, as illustrated in Fig.~\ref{fig:wallpaper} and Tab.~\ref{table:base_main}, providing valuable guidance for radar feature learning. 
Therefore, we propose a fusion-based KD method to extract informative cues from a LiDAR-camera fusion network (Teacher) and transfer them to the radar-based baseline R-SSC-RS and radar-camera fusion network, resulting in our approaches R-LiCROcc and RC-LiCROcc.

We utilize the same architecture for both LiDAR-camera and radar-camera fusion networks. For the fusion-based KD module, we combine Cross-Model Residual Distillation (CMRD), BEV Relation Distillation (BRD), and Predictive Distribution Distillation (PPD) to hierarchically compel the student model to learn the feature representations and distributions of the teacher model. Through this proposed fusion-based KD module, our LiCROcc with radar alone (R-LiCROcc) achieves comparable results against camera-based methods. Additionally, LiCROcc approaches the performance of LiDAR-based methods by incorporating both radar and camera inputs (RC-LiCROcc) while maintaining robustness in adverse weather conditions and night vision capabilities.

To summarize, the main contributions are as follows:
\begin{itemize}


    \item  We aim to improve radar for semantic scene completion while preserving real-world practicality, leveraging radar’s resilience to various weather conditions. We also establish radar-based benchmarks from LiDAR-based approaches, fostering radar-based SSC research, and consider a camera-radar fusion network for enhanced performance.

    \item  We resent a new framework, LiCROcc, which combines CMRD, BRD, and PPD modules to hierarchically force the student model to learn the feature representations and distributions of the teacher model. 
    
    \item  Extensive experiments on the large-scale nuScenes-Occupancy~\cite{wang2023openoccupancy} demonstrate the effectiveness of our proposed approaches. 
    %
    
\end{itemize}

\section{RELATED WORK}
\label{sec:related work}
\subsection{3D Semantic Scene Completion}

\textbf{LiDAR/Camera-based methods}. LiDAR-based methods \cite{zou2021up, roldao2020lmscnet, mei2023ssc, zuo2023pointocc, xia2023scpnet, yang2021semantic, cao2023pasco} use LiDARs for precise 3D semantic occupancy prediction. 
SSCNet \cite{song2017semantic} and UDNet \cite{zou2021up} utilize 3D U-Nets but face computation overhead from empty voxels. LMSCNet \cite{roldao2020lmscnet} and SGCNet \cite{zhang2018efficient} improve efficiency with 2D CNNs and spatial group convolutions, respectively. 
Advanced methods focus on multi-view fusion \cite{cheng2021s3cnet}, local implicit functions \cite{rist2021semantic}, knowledge distillation \cite{xia2023scpnet}, and BEV representation \cite{yang2021semantic, mei2023ssc}. 
Recently, Pasco \cite{cao2023pasco} further extends the SSC task with instance-level information to produce a richer 3D scene understanding.
Camera-based methods \cite{huang2023tri, mei2023camera,  zhang2023occformer, yu2023flashocc,  wang2023panoocc, cao2022monoscene, hou2024fastocc} have become popular due to their rich visual cues and cost-effectiveness. Many methods have explored effective 3D scene representation learning for outdoor surrounding SSC. For instance, TPVFormer \cite{huang2023tri} introduces a tri-perspective view for detailed 3D structure representation. OccFormer \cite{zhang2023occformer} utilizes transformers for multi-scale voxel features. PanoOcc \cite{wang2023panoocc} achieves a unified occupancy representation for comprehensive 3D scene understanding.

\textbf{Multi-modal methods} \cite{pan2024co, wang2023openoccupancy, ming2024occfusion} combines multi-source sensor data (e.g., images, LiDARs, and radars) to perform robust outdoor SSC. OpenOccupancy \cite{wang2023openoccupancy} provides a surrounding benchmark and establishes camera-based,
LiDAR-based and LiDAR-camera baselines. 
Recently, radar perception \cite{zhao2024crkd, kim2023rcm, zhou2023bridging, kim2023craft, nabati2021centerfusion} has garnered wide attention in multi-modal 3D detection task due to its cost-effectiveness, robustness against adverse weather conditions, and ability to detect distant objects. However, there are only a few works \cite{ming2024occfusion} to incorporate radar for outdoor SSC tasks. For instance, the recent OccFusion \cite{ming2024occfusion} devises a sensor fusion framework to integrate features from LiDARs, surround view images, and radars for robust and accurate SSC. We focus more on the distillation design of LC modes distilled to RC or R than on the direct fusion of multiple modes.

\begin{figure*}[htbp]
    \vspace{3pt}
    \captionsetup{font={small}}
    \centering
    \includegraphics[width=\linewidth]{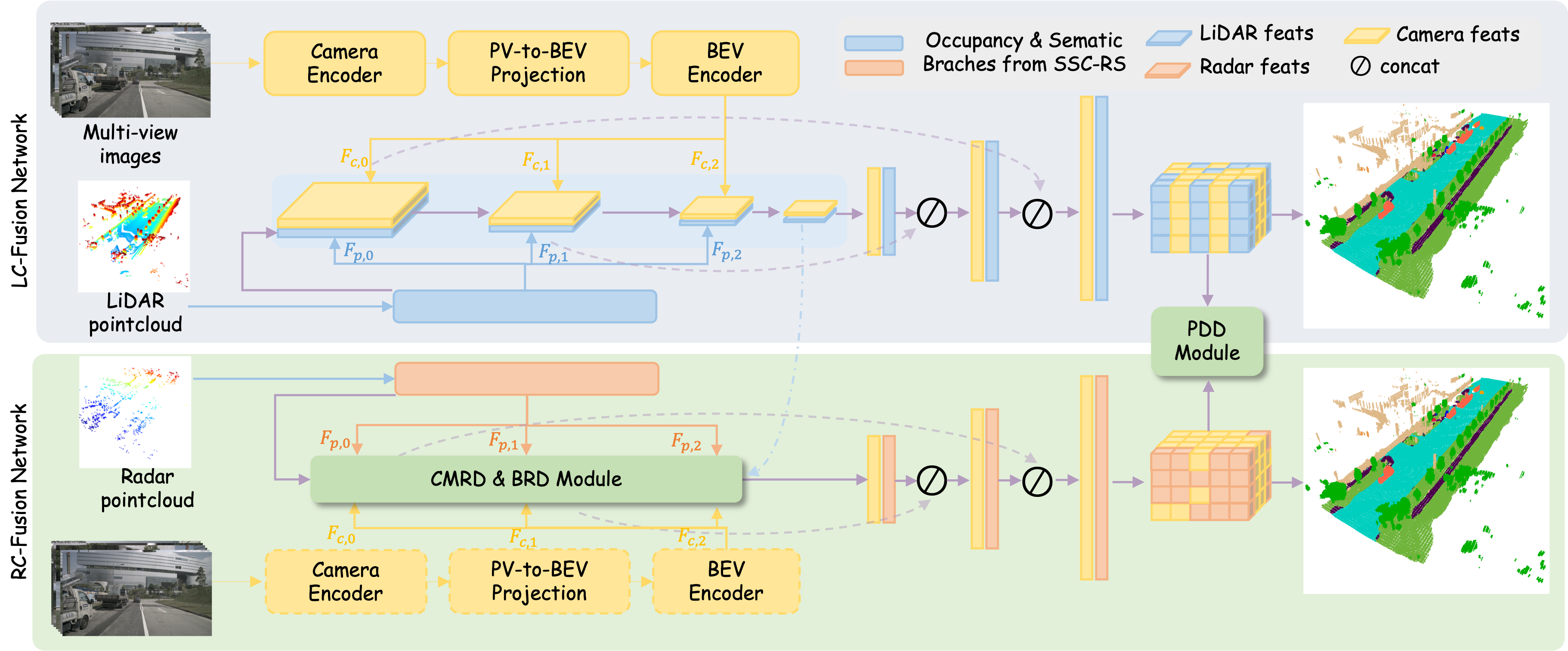}
    \caption{Overall framework of our {\methodname}. We designed a base framework for point cloud and image fusion, where the models serve as the teacher (LC-Fusion Network) and student (RC-Fusion Network), respectively. We unified both fusion processes under BEV space to reduce computational cost. Additionally, we designed three novel distillation losses (CMRD, BRD and PDD) to achieve effective cross-modal KD. Inference is performed using only the RC fusion Network, where camera input is optional. For the detailed structure of CMRD and BRD, please refer to Fig.~\ref{fig:KD}.
}
    \label{fig:pipeline}
\vspace{-10pt}
\end{figure*}

\subsection{Knowledge Distillation in Semantic Scene Completion}
The concept of Knowledge Distillation (KD) was initially proposed for model compression and performance improvement in image classification tasks \cite{hinton2015distilling}. It has since been extended to other fields, such as 3D object detection \cite{zhao2024crkd, zhou2023unidistill, chen2022bevdistill}, 3D segmentation \cite{hou2022point, yan20222dpass}, and semantic scene completion \cite{xia2023scpnet, zheng2024monoocc}.

In SSC tasks, SCPNet \cite{xia2023scpnet} introduces Dense-to-Sparse Knowledge Distillation (DSKD) to transfer dense, relation-based semantic knowledge from a multi-LiDAR teacher to a single-LiDAR student, significantly boosting the representation learning of the student model. Similarly, MonoOcc \cite{zheng2024monoocc} leverages knowledge distillation to transfer temporal information and visual cues to a monocular semantic occupancy framework. Different from the above methods that apply knowledge distillation in the same modality, our LiCROcc proposes fusion-based cross-modal knowledge distillation in a shared BEV space for outdoor SSC, passing informative cues from the LiDAR-camera features to the radar-camera features for performance improvement.
\section{METHODOLOGY}
\label{sec:method}

\subsection{Overview}
As mentioned above, we construct the radar-based baseline and design the radar-camera fusion network (the bottom part of Fig.~\ref{fig:pipeline}) to boost the baseline's performance. To leverage the guidance of detailed geometric structure and point representation in LiDAR-camera fusion, we further utilize the fusion-based KD (Section \ref{Sec:KD}) to transfer the knowledge from the LiDAR-camera fusion network (the top part of Fig.~\ref{fig:pipeline}) to the radar-based baseline and radar-camera fusion network. We employ the same architecture, i.e., the multi-modal fusion network (Section \ref{Sec:fusion}), to establish the above two fusion networks. 

\subsection{Multi-modal Fusion Network} \label{Sec:fusion}
The multi-modal fusion network mainly consists of the image branch for extracting image features, the point branch for encoding LiDAR/radar points, and the multi-modal BEV fusion network for effectively and efficiently integrating point and image features hierarchically.

\textbf{Image branch.} Following FlashOcc \cite{yu2023flashocc}, we propose to project surrounding image features to BEV space for subsequent processing, alleviating the memory overhead while maintaining the high accuracy of occupancy prediction. As shown in Fig.~\ref{fig:pipeline}, the image branch mainly consists of three components: the camera encoder for image features, the PV-to-BEV projection layer for BEV representation of the 3D scene, and the BEV encoder for hierarchical BEV features $ (\boldsymbol{F}_{c,0}, \boldsymbol{F}_{c,1}, \boldsymbol{F}_{c,2}$, where $\boldsymbol{F}_{c,i}\in \mathbb{R}^{C_i,H_i,W_i}) $ that contain the rich semantic context. The extracted multi-scale BEV features are fed into the BEV fusion model to interact with the point features, which will be elaborated on below.

\textbf{Point branch.} Without losing generality, we adopt the recent BEV-based SSC-RS \cite{mei2023ssc} as our point branch.
This branch uses two independent branches for semantic and geometric encoding. 
The BEV fusion network with an ARF module~\cite{mei2023ssc} aggregates features from these branches, resulting in the final semantic scene completion.
Due to its disentangled design, SSC-RS is lightweight and has strong representation ability, making it very suitable for use as the point branch. The point branch takes the LiDAR/radar point cloud $P$ and outputs the multi-scale BEV features $ (\boldsymbol{F}_{p,0}, \boldsymbol{F}_{p,1}, \boldsymbol{F}_{p,2}$, where $\boldsymbol{F}_{p,i}\in \mathbb{R}^{C_i,H_i,W_i}))$. For LiDAR point cloud, the $\boldsymbol{P}\in \mathbb{R}^{N\times3}$ is in the range of $[R_x, R_y, R_z]$. Moreover for the radar point cloud, $\boldsymbol{P}\in \mathbb{R}^{N\times6}$ is the concatenation of the $xyz$ coordinates, radar cross section $\sigma$, and the $xy$ velocities compensated by the ego-motion.

\textbf{Multi-modal BEV Fusion network.} 
\label{method:fusion model}
Due to the computational burden associated with 3D convolutions for dense feature fusion, we introduce a multi-modal bev fusion network, drawing inspiration from BEV perception tasks. This network efficiently combines semantically rich visual BEV representations $ (\boldsymbol{F}_{c,0}, \boldsymbol{F}_{c,1}, \boldsymbol{F}_{c,2}) $, geometrically informative LiDAR features or weather-resistant radar features. To streamline the fusion process, we unify LiDAR or radar point cloud features with $ (\boldsymbol{F}_{p,0}, \boldsymbol{F}_{p,1}, \boldsymbol{F}_{p,2})$.
Similarly to \cite{mei2023ssc}, our BEV fusion network employsa 2D convolutional U-Net architecture. Each residual block reduces the input feature resolution by a factor of 2 to maintain consistency with semantic/complementary features. Before each subsequent block, we integrate the previous stage's $\boldsymbol{F}_{b,i-1}$ with the current stage's $\boldsymbol{F}_{p,i}$ using ARF~\cite{mei2023ssc} to obtain $\boldsymbol{F}_{b_i}$, and then the scaled $\boldsymbol{F}_{c,i}$ is fused to $\boldsymbol{F}_{b,i}$ by addition. The decoder upsamples the encoder's compressed features three times by a factor of two by skipping connections. The final decoder convolution generates the SSC prediction $ \boldsymbol{Y} \in \mathbb{R}^{((C_n+1)\cdot Z) \times H \times W} $, where $ C_n $ denotes the number of semantic classes. To represent the voxel-wise semantic occupancy probabilities, $ \boldsymbol{Y} $ is reshaped into $( ({C_n+1})\times{Z} \times H \times W )$.

To train the proposed fusion model, cross-entropy loss $\mathcal{L}_{ce}$ is used to optimize the network. In addition, following~\cite{cao2022monoscene}, we also utilize affinity loss $\mathcal{L}_{\rm{scal}}^{\rm{geo}}$ and $\mathcal{L}_{\rm{scal}}^{\rm{sem}}$ to optimize the metrics in the scene and the class (\ie, geometric IoU, and semantic mIoU). Therefore, the BEV fusion loss function can be derived as:
\begin{equation}
    \mathcal{L}_{\rm{bev}} = \mathcal{L}_{\rm{ce}} + \mathcal{L}_{\rm{scal}}^{\rm{geo}} + \mathcal{L}_{\rm{scal}}^{\rm{sem}},
\end{equation}

\begin{figure}[t]
    \captionsetup{font={small}}
    \centering
    \includegraphics[width=\linewidth]{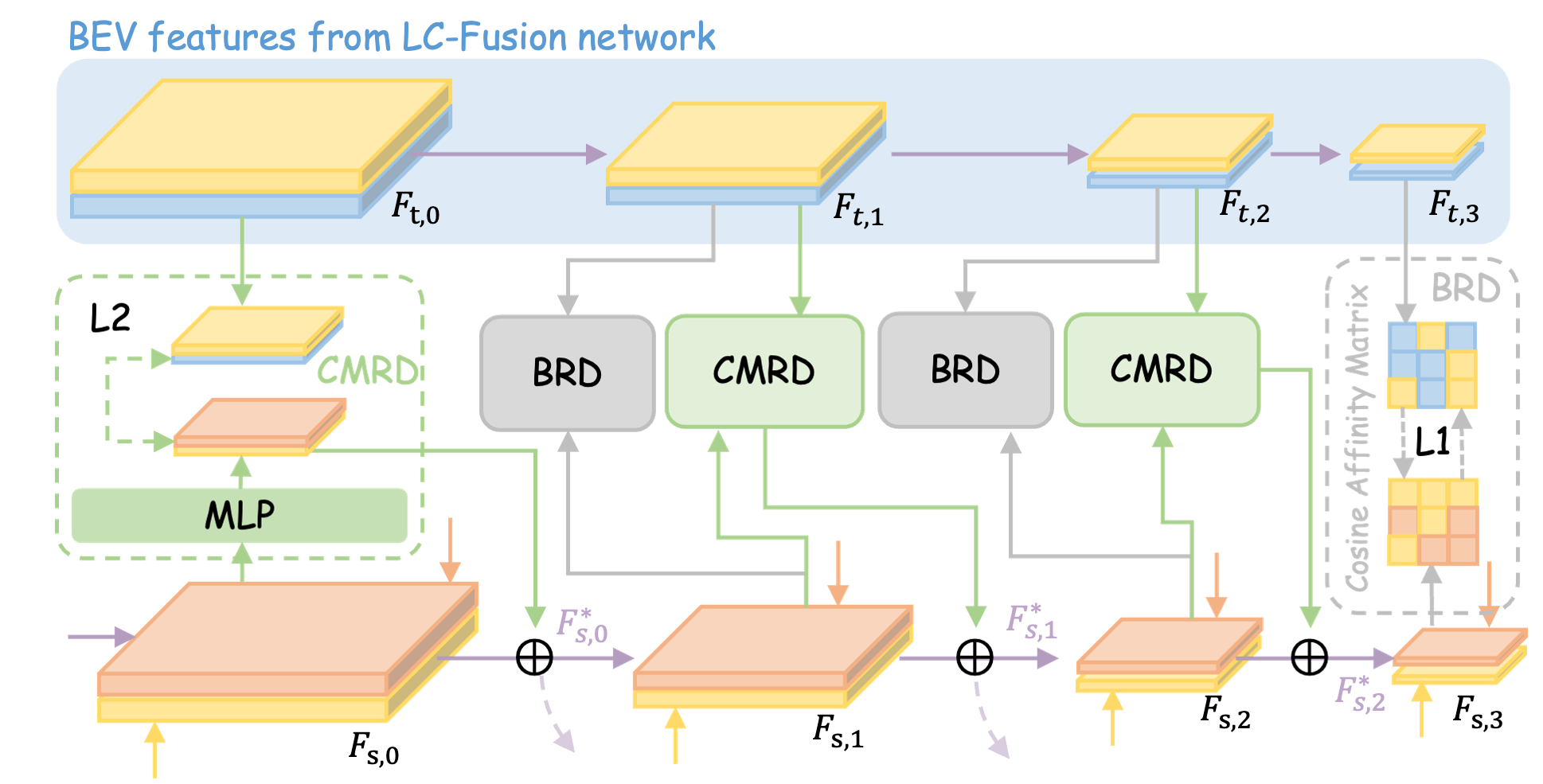}
    \caption{Detailed illustration of the use of CMRD and BRD on BEV features. The feature with the blue background is a BEV feature duplicated from the teacher model. The details of the loss computation are illustrated in the dashed box. The $\oplus$ in the figure denotes the summation of features.
    }
    \label{fig:KD}
\vspace{-15pt}
\end{figure}

\subsection{Fusion-based Knowledge Distillation Module} \label{Sec:KD}
As shown in Fig.~\ref{fig:pipeline}, our teacher and student model share the same network structure. Thanks to the fact that both the fusion and distillation processes are under BEV, the image branching of the student model is optional.
 In this section, we use $(\boldsymbol{F}_{s,0}, \boldsymbol{F}_{s,1}, \boldsymbol{F}_{s,2}, \boldsymbol{F}_{s,3})$ to represent the four BEV features of the student model, and $(\boldsymbol{F}_{t,0}, \boldsymbol{F}_{t,1}, \boldsymbol{F}_{t,2}, \boldsymbol{F}_{t,3})$ to represent the corresponding features of the teacher model (illustrated in light blue in Fig.~\ref{fig:pipeline} and Fig.~\ref{fig:KD}).

\subsubsection{Cross-Model Residual Distillation}
Camera and LiDAR fusion features contain rich semantic and geometric information. Compared to them, radar points are much sparser. The semantic information of radar is mainly derived from velocity measurements. 
Observing this gap, we believe the standard approach of directly forcing radar features to mimic multi-modal features may not work well \cite{zhao2024crkd}, so we design a Cross-Model Residual Distillation module. Specifically, we use Eq.~\ref{eq:MLP} to project student features $\boldsymbol{F}_{s}$ onto a hidden space $\boldsymbol{F}'_{s}$ with the same dimensions. 
In that space, we bring the features of the student and teacher closer to each other by minimizing the cosine similarity (calculated by Eq.~\ref{eq:cmrd}) between them and eventually add $\boldsymbol{F}'_{s}$ back to the original $F_{s}$ as an information complement, which avoids disturbing the intrinsic logic of the radar features compared to direct feature mimicry approaches.
Moreover, radar possesses a distinct advantage over cameras and LiDAR regarding weather resilience and observation range. Our goal is for the student model to learn from the teacher’s strengths while maintaining its unique characteristics rather than simply copying the teacher. Based on the above, we utilize ARF to dynamically calculate the weights for integrating $\boldsymbol{F}'_{s}$ with $\boldsymbol{F}_{s}$. The procedure for feature transfer is outlined as follows:
\begin{align}
    \boldsymbol{F}'_{s} &= MLP(\boldsymbol{F}_{s}) \\
    \boldsymbol{F}_{s}^* &= \boldsymbol{F}_{s}+\boldsymbol{F}'_{s}\times ARF(\boldsymbol{F}'_{s})
    \label{eq:MLP}
\end{align}
Assuming that $\boldsymbol{f}_{(u,v)}$ is a feature indexed as $(u, v)$ on the feature map $\boldsymbol{F}$, the CMRD loss $\mathcal L_{cmrd}$ is formed as follows:
\begin{equation}
    \mathcal L_{cmrd} = \dfrac{1}{H\times W}\sum_u^H{ \sum_v^W {M_{u,v} \dfrac{{\boldsymbol{f}'}_{s,(u,v)}^\top \boldsymbol{f}_{t,(u,v)}}{\|{\boldsymbol{f}'}_{s,(u,v)}\|_2 \|\boldsymbol{f}_{t,(u,v)}\|_2}}}
    \label{eq:cmrd}
\end{equation}
where $M_{u,v}$ =1 if there are labels on that pillar that are non-empty and non-noise, and $M_{u,v}=0$ otherwise. In other words, we constrain the feature similarity only on occupied locations. As shown in Fig.~\ref{fig:KD}, we computed $\mathcal L_{cmrd}$ for $\boldsymbol{F}_{s,i}$ and $\boldsymbol{F}_{t,i}$ for $i=0,1,2$, where the green dashed box shows the details of the computed loss.

\subsubsection{BEV Relation Distillation}
%
We employ the CMRD for feature-level cross-modal distillation, which maintains the integrity of the student features while enriching them with additional teacher information. This section introduces a mechanism designed to uphold the consistency of scene-level geometric relationships. To achieve this, we employ a cosine similarity-based affinity matrix to compare teacher feature maps $\boldsymbol{F}_{t}$  and student feature maps $\boldsymbol{F}_{s}$. Initially, the tensors $ \boldsymbol{F}_{s} $ and $ \boldsymbol{F}_{t} $ are defined in the space $\mathbb{R}^{C \times H \times W}$. We then transform these tensors into matrices with dimensions $C \times (H \cdot W)$. 
The affinity matrix is computed using the following calculation:
\begin{align}
    C_{u,v} = \dfrac{{\boldsymbol{f}_{u}}^\top \boldsymbol{f}_{v}}{\|{\boldsymbol{f}_{u}}\|_{2} \|\boldsymbol{f}_{v}\|_2}, (u,v\in \left\{1,2,\cdots,K= H \times W\right\}),
\end{align}
where $C_{u,v}$ dentoes the cosine similarity at each element $(u, v)$ of the affinity matrix, $\boldsymbol{f}_{u}$ denotes the $u$-th feature in the feature map $\boldsymbol{F}$. To assess the scene-level information gap between the student and teacher model, we compute the L1 norm between their affinity matrices. The BRD loss is then defined as follows:
\begin{align}
    \mathcal L_{brd} = \dfrac{1}{K \times K} \sum_{u=1}^K \sum^K_{v=1} \|C^S_{u,v} - C^T_{u,v}\|_1,
\end{align}

\begin{table*}
    \vspace{3pt}
    \captionsetup{font={small}}
	\setlength{\tabcolsep}{0.0035\linewidth}
	\newcommand{\classfreq}[1]{{~\tiny(\semkitfreq{#1}\%)}}  %
	\centering
   \resizebox{1\linewidth}{!}{
	\begin{tabular}{l|c c | c c | c c c c c c c c c c c c c c c c}
 
		\toprule
		Method
		& \makecell[c]{Input}
		& \makecell[c]{Surround}
		& \makecell[c]{IoU}
            & \makecell[c]{mIoU}
		& \rotatebox{90}{\textcolor{barrier}{$\blacksquare$} barrier} 
		& \rotatebox{90}{\textcolor{bicycle}{$\blacksquare$} bicycle}
		& \rotatebox{90}{\textcolor{bus}{$\blacksquare$} bus} 
		& \rotatebox{90}{\textcolor{car}{$\blacksquare$} car} 
		& \rotatebox{90}{\textcolor{const. veh.}{$\blacksquare$} const. veh.} 
		& \rotatebox{90}{\textcolor{motorcycle}{$\blacksquare$} motorcycle} 
		& \rotatebox{90}{\textcolor{pedestrian}{$\blacksquare$} pedestrian} 
		& \rotatebox{90}{\textcolor{traffic cone}{$\blacksquare$} traffic cone} 
		& \rotatebox{90}{\textcolor{trailer}{$\blacksquare$} trailer} 
		& \rotatebox{90}{\textcolor{truck}{$\blacksquare$} truck} 
		& \rotatebox{90}{\textcolor{drive. suf.}{$\blacksquare$} drive. suf.} 
		& \rotatebox{90}{\textcolor{other flat}{$\blacksquare$} other flat} 
		& \rotatebox{90}{\textcolor{sidewalk}{$\blacksquare$} sidewalk} 
		& \rotatebox{90}{\textcolor{terrain}{$\blacksquare$} terrain} 
		& \rotatebox{90}{\textcolor{manmade}{$\blacksquare$} manmade} 
		& \rotatebox{90}{\textcolor{vegetation}{$\blacksquare$} vegetation} \\
		\midrule
		MonoScene~\cite{cao2022monoscene} & C & \xmark  & 18.4 & 6.9 & 7.1  & 3.9  &  9.3 &  7.2 & 5.6  & 3.0  &  5.9& 4.4& 4.9 & 4.2 & 14.9 & 6.3  & 7.9 & 7.4  & 10.0 & 7.6 \\
  
  		TPVFormer~\cite{huang2023tri} &C &  \cmark& 15.3 &  7.8 & 9.3  & 4.1  &  11.3 &  10.1 & 5.2  & 4.3  & 5.9 & 5.3&  6.8& 6.5 & 13.6 & 9.0  & 8.3 & 8.0  & 9.2 & 8.2 \\
    
            3DSketch~\cite{chen20203d} &  C\&D & \xmark& 25.6 & 10.7  & 12.0 &  5.1 &  10.7 &  12.4 & 6.5  & 4.0  & 5.0 & 6.3&  8.0&  7.2& 21.8 &  14.8 & 13.0 &  11.8 & 12.0 & 21.2 \\
            
            AICNet~\cite{li2020anisotropic} & C\&D   &  \xmark& 23.8 & 10.6  & 11.5  & 4.0  & 11.8  & 12.3&  5.1 & 3.8  & 6.2  & 6.0 & 8.2&  7.5&  24.1 & 13.0 & 12.8  & 11.5 & 11.6  &  20.2\\

            LMSCNet~\cite{roldao2020lmscnet} & L &  \cmark& 27.3 & 11.5 & 12.4&  4.2 & 12.8  & 12.1  & 6.2  &  4.7 & 6.2 & 6.3&  8.8&  7.2& 24.2 & 12.3  & 16.6 & 14.1  & 13.9 & 22.2 \\
    
    		JS3C-Net~\cite{yan2021sparse} &L &  \cmark& 30.2  & 12.5 & 14.2 & 3.4  & 13.6  & 12.0  & 7.2  &  4.3 & 7.3 & 6.8&  9.2& 9.1 & 27.9 & 15.3  & 14.9 & 16.2  & 14.0 & {24.9} \\
      
            C-CONet~\cite{wang2023openoccupancy}  & C &  \cmark&20.1  & 12.8&13.2  & 8.1 &  15.4 &  17.2 & 6.3  & 11.2  & 10.0  &  8.3 & 4.7 & 12.1 & 31.4 & 18.8 & 18.7  & 16.3 & 4.8  &8.2  \\
            
            L-CONet~\cite{wang2023openoccupancy}  & L &  \cmark& {30.9}  & 15.8 &  17.5  & 5.2  & 13.3  & 18.1  & 7.8  & 5.4  & 9.6 & 5.6& 13.2 & 13.6 & {34.9} & {21.5}  & 22.4 & {21.7}  & 19.2 &23.5  \\
            
             PointOcc~\cite{zuo2023pointocc} &L & \cmark & 34.1 & 23.9 & 24.9 & 19.0& 20.9 & 25.7 & 13.4 & \textbf{25.6} & \textbf{30.6} & 17.9 & 16.7 & 21.2 & 36.5 & \textbf{25.6} & {25.7} & {24.9} & {24.8} & {29.0}  \\
             
            M-CONet~\cite{wang2023openoccupancy} & C\&L &   \cmark& 29.5 & {20.1} &  {23.3}  & {13.3}  & {21.2}  & {24.3}  & {15.3}  & {15.9}  & {18.0} & {13.3} & {15.3} &{ 20.7} & 33.2 & 21.0 & {22.5}  & 21.5 & {19.6}  & 23.2  \\

            Co-Occ~\cite{pan2024co} & C\&L &   \cmark &30.6&21.9&26.5&16.8&\textbf{22.3}&\textbf{27.0}&10.1&20.9&20.7 &14.5 &16.4 &21.6&\textbf{36.9}&23.5&2\textbf{5.5} &23.7 &20.5 &23.5  \\

            {LC-{Fusion}} (Ours) & C\&L & \cmark& \textbf{34.9} & \textbf{24.7} & \textbf{29.6} & \textbf{20.5} & 22.2 & 26.4 & \textbf{15.7} & 24.5 & 27.3 & \textbf{21.8} & \textbf{18.1}& \textbf{21.8} & 35.9 & 22.8 & 25.0 & \textbf{25.1}& \textbf{27.8} & \textbf{30.5} \\

            

		\midrule
            
            {R-CONet}~\cite{wang2023openoccupancy} & R & \cmark& 17.0 & 5.9 & 6.3 & 0.6 & 3.4 & 9.4 & 0.9 & 0.9 & 1.0 & 1.7 & 2.3 & 3.9 & 24.2 & 8.8 & 11.4 & 8.6 & 6.1 & 4.2  \\
            {R-SSC-RS}~\cite{mei2023ssc} & R & \cmark& 20.8 & 7.5 & 5.3 & 0.3 & 5.4 & 13.1 & 1.7 & 1.4 & 7.4 & 2.3 & 2.6 & 7.0 & 24.2 & 8.5 & 11.4 & 9.4 & 9.2 & 10.5  \\
            
            {PointOcc}~\cite{zuo2023pointocc} & R & \cmark& \textbf{21.9} & 7.4 & 4.9 & 0.8 & 5.7 & 13.1 & 1.6 & 2.1 & 6.1 & 1.6 & 2.5 & 5.9 & \textbf{26.4} & 8.1 & 11.7 & 8.7 & 9.9 & 10.0  \\
            
            {R-{\methodname}} (Ours) & R & \cmark& 21.3 & \textbf{8.5} & \textbf{8.1} & \textbf{0.9} & \textbf{6.4} & \textbf{13.6} & \textbf{2.3} & \textbf{2.7} & \textbf{7.9} & \textbf{2.7} & \textbf{3.3} & \textbf{7.6} & 24.3 & \textbf{11.2} & \textbf{13.1} & \textbf{10.7} & \textbf{10.6} & \textbf{11.0}   \\ 
		\midrule
            
            RC-CONet~\cite{wang2023openoccupancy} & C\&R & \cmark& 19.0 & 14.8 & 16.7 & 11.6 & 17.0 & 20.6 & 8.6 & 15.5 & 15.0 & 11.2 & 6.5 & 15.5 & 28.0 & 19.7 & 18.7 & 15.6 & 7.7 & 8.8  \\

             Co-Occ~\cite{pan2024co} & C\&R &   \cmark &24.2& 16.6 &18.6&\textbf{12.6}&\textbf{18.1}&\textbf{23.0}&6.4&\textbf{16.5}&15.2 &11.2 &7.0 &15.3&\textbf{34.3}&\textbf{21.9}&\textbf{23.0} &19.6 &10.3 &12.0  \\
            RC-{Fusion} (Ours) & C\&R & \cmark& 25.2 & 15.6 & 14.6 & 10.4 & 16.4 & 20.5 & 9.5 & 15.0 & 15.5 & 10.0 & 7.0 & 15.0 & 32.2 & 18.5 & 20.3 & 18.3 & 11.6 & 14.8  \\ 
            RC-{\methodname} (Ours) & C\&R & \cmark& \textbf{26.0} & \textbf{17.1} & \textbf{18.6} & 11.9 & 17.1 & 21.6 & \textbf{11.1} & 15.5 & \textbf{16.7} & \textbf{11.5} & \textbf{8.8} & \textbf{16.0} & 34.1 & 20.9 & 21.9 & \textbf{19.7} & \textbf{12.8} & \textbf{15.9}  \\

        \bottomrule
	\end{tabular}} \\

	\caption{Performance on nuScenes-Occupancy (validation set). We report the geometric metric IoU, semantic metric mIoU, and the IoU for each semantic class. The $C,D,L,R,M$ denotes \textit{camera, depth, LiDAR, radar} and \textit{multi-modal}. For \textit{Surround=}\cmark, the method directly predicts surrounding semantic occupancy with 360-degree inputs. Otherwise, the method produces the results of each camera view, and then concatenates them as surrounding outputs. We divide the form into three categories based on the modality of the inputs. Bold represents the best score.
}
	\label{table:base_main}
    \vspace{-15pt}
\end{table*}
where $\boldsymbol{C}^S, \boldsymbol{C}^T$ denote the affinity matrix of the student and teacher network bev feature maps, respectively. As shown in Fig.~\ref{fig:KD}, we computed $\mathcal L_{brd}$ for $\boldsymbol{F}_{s,i}$ and $\boldsymbol{F}_{t,i}$ for $i=0,2,3$, where the gray dashed box shows the details of the computed loss. To alleviate the computation burden, we resize all the BEV features of different scales to a smaller resolution and then compute $\mathcal L_{brd}$. 
It is worth noting that BRD has only been used to distill to the radar-camera fusion for the student model. 

\subsubsection{Predictive Distribution Distillation}
 Our work introduces a novel knowledge distillation method utilizing KL divergence. Specifically, we compute the KL divergence upon probabilities $\widetilde{\boldsymbol{Y}} = \rm{softmax}$$ (\boldsymbol{Y})$, where $\boldsymbol{Y}$ is introduced in Sec.~\ref{Sec:fusion}, predicted by teacher and student models. This measure captures the distribution discrepancy and is integrated into the distillation objective. By minimizing the KL divergence, the student model is encouraged to closely align its predictions with the teacher's, thereby enhancing its predictive capabilities. The PDD loss can be computed as follows:
 \begin{align}
     \mathcal L_{pdd} = \text{KL}(\widetilde{\boldsymbol{Y}}_S || \widetilde{\boldsymbol{Y}}_T),
 \end{align}
 where $\widetilde{\boldsymbol{Y}}_S$ is the predictive probability distribution of the student model and $\widetilde{\boldsymbol{Y}}_T$, the learning target, is the predictive probability distribution of the teacher model.

 \subsection{Overall Loss Functions}
During the training phase, we adopt a multi-task training strategy to guide the various components effectively. For 3D semantic scene completion, we utilize the $\mathcal{L}_{bev}$ loss. To enhance the distillation process, we combine three distillation components: $\mathcal{L}_{red}$, $\mathcal{L}_{brd}$, and $\mathcal{L}_{prdd}$. Additionally, we retain the semantic ($L_s$) and occupancy ($L_c$) losses from SSC-RS~\cite{mei2023ssc} to supervise feature point cloud extraction. The overall loss function is represented as follows:
\begin{align}
\begin{split}
    \mathcal{L} &= \lambda_1 \mathcal{L}_{bev}+ \lambda_2 \mathcal{L}_{cmrd} + \lambda_3     \mathcal{L}_{brd} \\ 
    &+ \lambda_4 \mathcal{L}_{pdd} + \lambda_5(\mathcal{L}_c +\mathcal{L}_s)
\end{split}
\end{align}
where $\lambda_1,\lambda_2,\lambda_3,\lambda_4$ and $\lambda_5$ are hyper-parameters.
\section{EXPERIMENTS} 
In this section, we elaborate on the evaluation datasets and metrics (Sec.~\ref{exp:Dataset and Metrics}), implementation details (Sec.~\ref{exp:Implement Details}), and performance comparisons with state-of-the-art methods (Sec.~\ref{exp:Quantitative Results}). Additionally, we conduct ablation studies to demonstrate the effectiveness of the proposed fusion module (Sec.~\ref{exp:Effect of Fusion Module}) and distillation module (Sec.~\ref{exp:Effect of Distillation}). Finally, we provide experiments to ablate the impact of the observation distance~(Sec.~\ref{exp:Visual Field Benefits from Radar}) and the unique benefits of radar for semantic scene completion task (Sec.~\ref{exp:Weather Robustness from Radar}).

\subsection{Dataset and Metrics}
\label{exp:Dataset and Metrics}
\textbf{Datasets.} We evaluate our approach on the nuScenes-Occupancy~\cite{wang2023openoccupancy} benchmark, which comprises 850 scenes (700 for training and 150 for validation) with 34,000 keyframes at the 3D volume with the size of $40 \times 512 \times 512$. The dataset covers an area of -51.2 to 51.2 meters in the $xy$ plane and -5 to 3 meters in the $z$-axis. It provides the voxel annotations with 17 categories and a $0.2m \times 0.2m \times 0.2m$ resolution.

\textbf{Metrics.} Following \cite{song2017semantic}, we employ Intersection-over-Union (IoU) for scene completion (excluding semantics), and the mean Intersection-over-Union (mIoU) for semantic scene completion  (no ``noise" class), as the validation metrics.
\subsection{Implementation Details}
\label{exp:Implement Details}
For LiDAR inputs, we concatenate 10 LiDAR sweeps as a keyframe similar to \cite{wang2023openoccupancy, zuo2023pointocc}. We leverage a pre-trained ResNet50~\cite{he2016deep} on ImageNet~\cite{deng2009imagenet} as the image backbone to process the camera images with the input resolution of $256 \times 704$. 
During the training, we project the point cloud onto the camera view to provide depth supervision of the LSS~\cite{philion2020lift}. 
For radar input, we adopt the preprocessing process in CRN~\cite{kim2023crn}, using radar scans stitched from 5 radar sensors of the car.
For data augmentation, we randomly apply horizontal and vertical flips and cropping to the images. The point clouds are augmented by the random flipping on the $x$-axis and $y$-axis.
We employ the AdamW optimizer with a weight decay of 0.01 and an initial learning rate of 2e-4. We use the cosine learning rate scheduler with linear warming up in the first 500 iterations.
All experiments are conducted on 8 NVIDIA A100 GPUs with a total batch size 32 for 24 epochs.

 \begin{table}[t]
\vspace{3pt}
\captionsetup{font={small}}
  \centering
  \resizebox{\linewidth}{!}{
  \begin{tabular}{c|c|cc||c|c|cc}
    \toprule
    Input & stages  & IoU$\uparrow$ & mIoU$\uparrow$ & Input & stages  & IoU$\uparrow$ & mIoU$\uparrow$ \\
    \midrule
    L & 0 & 34.4 & 22.1   &  R & 0 & 20.8 & 7.5\\ 
    \midrule
    \rowcolor{black!2}
    \cellcolor{white} ~ & 1 &   34.0 & 23.4   &  \cellcolor{white} ~ & 1 &   24.7 & 15.0 \\ 
    \rowcolor{black!6}
    \cellcolor{white} ~ & 2 &  34.2 & 23.7  &  \cellcolor{white} ~ & 2 &  24.9 & 15.4 \\ 
    \rowcolor{black!10}
    \cellcolor{white} \multirow{-3}*{C\&L} & 3  &  \textbf{34.9} & \textbf{24.7}  &  \cellcolor{white} \multirow{-3}*{C\&R} & 3  &  \textbf{25.2} & \textbf{15.6}\\
    \bottomrule
  \end{tabular}
  }
  \caption{Ablation on fusion stages.}
  \label{tab:fusion_stage}
    \vspace{-15pt}
\end{table}

\subsection{Quantitative Results}
\label{exp:Quantitative Results}
Tab.~\ref{table:base_main} shows the comparison results of our method against the state-of-the-art methods on the nuScenes-Occupancy \cite{wang2023openoccupancy} benchmark. Compared with all previous methods, our {\methodname} achieves the best performance under the same configuration. For example, our LiDAR-camera fusion model {LC-Fusion} shows significant improvements, with a 23\% increase in mIoU and an 18\% increase in IoU compared to the baseline (M-CONet~\cite{wang2023openoccupancy}). Meanwhile, {LC-Fusion} achieves the improvements of 3.3\% and 2.3\% in terms of mIoU and IoU scores over PointOCC~\cite{zuo2023pointocc}, which emphasizes the effectiveness of our proposed multi-modal BEV fusion.

\begin{figure*}[htbp]
    \captionsetup{font={small}}
    \centering
    \includegraphics[width=\linewidth]{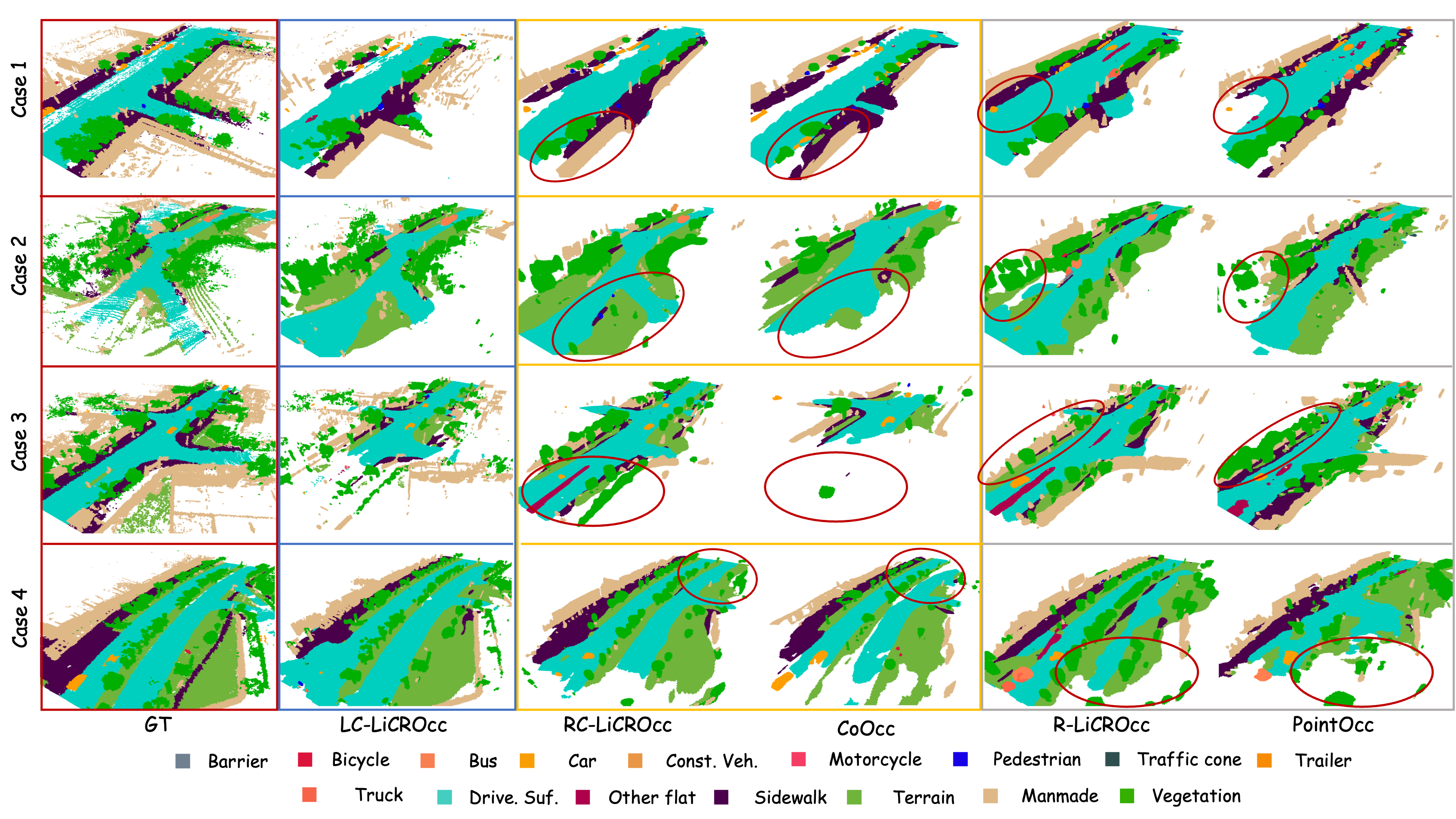}
    \caption{Visual comparison of our LiCROcc with baseline methods on OppenOccupancuy benchmark. Our methods stand out by offering a more comprehensive representation of the scene and more accurate segmentation boundaries compared to CoOcc and PointOcc. Surprisingly, the R-LiCROcc can segment feasible regions and obstacles even with only two thousand radar points as input.
}
    \label{fig:visualization}
\end{figure*}

To comprehensively evaluate the effectiveness of our proposed method, we have modified several existing LiDAR-based and multi-modal methods (CONet~\cite{wang2023openoccupancy}, SSC-RS~\cite{mei2023ssc}, PointOcc~\cite{zuo2023pointocc}, and CoOCC~\cite{pan2024co}), to accommodate radar input, serving as our comparisons. 
As shown in the second part of Tab.~\ref{table:base_main}, 
our {R-\methodname} outperforms the second-best one (PointOcc) and the baseline (R-SSC-RS) by 13.3\% on the mIoU scores, demonstrating the effectiveness of our proposed fusion-based knowledge distillation. We find the IoU score of our {R-\methodname} is slightly lower than PointOcc. We explain that this can be attributed to the fact that PointOcc projects features on three planes and uses a larger model, which may be more beneficial for occupancy prediction.
For radar-camera fusion, we take CONet and CoOCC as our baselines. The results are presented in the third part of Tab.~\ref{table:base_main} and show that our radar-camera fusion version RC-Fusion has achieved comparable performance to these baselines. The proposed fusion-based knowledge distillation further boosts the performance by 1.5 and 0.8 in terms of mIoU and IoU. We also provide the visualization in Fig.~\ref{fig:visualization} illustrates that our {RC-\methodname} and {R-\methodname} enable more complete scene completion and precise object segmentation.

\subsection{Ablation Studies}
\label{exp:Ablation Studies}
We conduct a series of experiments to validate the proposed module and the potential of radar as a sensor for semantic scene completion tasks. All experiments are conducted under the same training configuration and evaluated according to nuScene-Occupancy \cite{wang2023openoccupancy} validation dataset.
\subsubsection{Effect of Fusion Module}
\label{exp:Effect of Fusion Module}
We investigate the impact of different fusion stages in the multi-modal BEV fusion network presented in \ref{Sec:fusion}. The corresponding results are shown in Tab.~\ref{tab:fusion_stage}. ``Stages=0" means only using the point cloud as input, which serves as the point-based baseline. From Tab.~\ref{tab:fusion_stage}, we can see that the multi-stage fusion strategy fuses BEV features at different scales and effectively improves the accuracy of the semantic scene completion.

\subsubsection{Effect of Distillation Module}
\begin{table}[t]
\vspace{3pt}
\captionsetup{font={small}}
  \centering
  \resizebox{0.85\linewidth}{!}{
  \begin{tabular}{c|ccc|cc}
    \toprule
    Model & CMRD & BRD &  PDD & IoU$\uparrow$ & mIoU$\uparrow$ \\
    \midrule
    ~ & & & &  20.8 & 7.5 \\  
    \rowcolor{black!2}
    \cellcolor{white} ~ & \checkmark & & &  \textbf{21.7} & 7.9 \\
    \rowcolor{black!6}
    \cellcolor{white} \multirow{-3}*{R-LiCROcc} & \checkmark &  & \checkmark &  21.3 & \textbf{8.5} \\
    \midrule
    ~ & & & &  25.2 & 15.6 \\  
    \rowcolor{black!2}
    \cellcolor{white} ~ & \checkmark & & &  25.3 & 16.1 \\
    \rowcolor{black!6}
    \cellcolor{white} ~ & \checkmark & \checkmark & &  25.6 & 16.3 \\
    \rowcolor{black!10}
    \cellcolor{white} \multirow{-4}*{RC-LiCROcc} & \checkmark & \checkmark & \checkmark &  \textbf{26.0} & \textbf{17.1}\\
    \bottomrule
  \end{tabular}
  }
  \caption{Ablation on distillation modules.}
  \label{tab:distillation}
    \vspace{-15pt}
\end{table}
\label{exp:Effect of Distillation}
In this section, we delve into the individual contributions of different distillation components of our proposed fusion-based knowledge distillation. Detailed results are illustrated in Tab.~\ref{tab:distillation}. We systematically evaluated the impact of these modules by incorporating them into two distinct configurations: R-LiCROcc (the first part in Tab.~\ref{tab:distillation}) and RC-LiCROcc (the second part in Tab.~\ref{tab:distillation}). Each configuration starts with a baseline model with any distillation module, to which we sequentially add our distillation modules to evaluate their efficacy.

Results in both parts of Tab. \ref{tab:distillation} show that CMRD, BRD, and PDD components significantly enhance the performance. Among these, PDD stands out with the most substantial boost, registering a 7.6\% mIoU improvement for {R-\methodname} and a 4.9\% mIoU improvement for {RC-\methodname}, underscoring its crucial role in cross-modal knowledge distillation. 

\subsubsection{Visual Field Benefits from KD}
\label{exp:Visual Field Benefits from Radar}
Radar's inherent ability to penetrate objects and bypass foreground obstacles enables it to provide a wider field of view than LiDAR and camera sensors. However, the sparsity of radar point clouds increases with distance, which is particularly unfavorable for SSC, as shown in rows 2 and 3 of Tab.~\ref{tab:range}.

In order to further analyze the improvements brought by {\methodname}, we conducted a statistical analysis to evaluate the system's effectiveness for semantic scene completion across various distance ranges, as detailed in Tab.~\ref{tab:range}. We measure the IoUs and mIoUs of the teacher model, the student model, and the R-{\methodname} at $[0\text{m}, 20\text{m}]$, $[20\text{m}, 30\text{m}]$, and $[30\text{m}, 50\text{m}]$ for semantic scene complementation, respectively.
Tab.\ref{tab:range} reveals that knowledge distillation (KD) significantly enhances the student model's performance, particularly in the short-range area. Interestingly, we found that when performing KD from LiDAR-camera fusion to radar-based models, the improvement in mIoU scores for the long-range area is much smaller than that for short- and medium-range areas. This observation suggests that the LiDAR-camera fusion loses its advantage as distance increases due to its relatively shorter visual range. 
It is worth noting that both the teacher and student models exhibit severe performance degradation in the long-range area, especially in the mIoU score. For example, the teacher model outperforms RC-{\methodname} by 10.96 mIoU within 20m. However, this advantage sharply drops to 3.76 points (almost 65\% decrease) in the $[30\text{m}, 50\text{m}]$ range.

%
\begin{table*}[t]
\captionsetup{font={small}}
    \centering
    \resizebox{0.9\linewidth}{!}{
     \begin{tabular}{c|c|ccc|ccc}
        \toprule
        & & \multicolumn{3}{c|}{IoU$\uparrow$} & \multicolumn{3}{c}{mIoU$\uparrow$}         \\
        \multirow{-2}{*}{Method} & \multirow{-2}{*}{Modality} & $[0\text{m}, 20\text{m}]$ & $[20\text{m}, 30\text{m}]$ & $[30\text{m}, 50\text{m}]$ & $[0\text{m}, 20\text{m}]$ & $[20\text{m}, 30\text{m}]$ & $[30\text{m}, 50\text{m}]$ \\ 
        \midrule
        LC-Fusion (Teacher) & L+C & 49.0 & 23.69 & 12.01 & 34.62 & 14.85 & 4.98\\
        \midrule
        R-SSC-RS & R & 28.12 & 7.07 & 0.91 & 9.38 & 2.25 & 0.39 \\
        \rowcolor{black!10}
        \cellcolor{white} 
        \textbf{R-\methodname} (Student) & R & 27.87(-0.25) & 9.56(+2.49) & 0.81(-0.1) & 10.44(+1.06) & 3.26(+1.01) & 0.81(+0.42) \\
        \midrule
        RC-Fusion & C+R & 35.79 & 10.56 & 1.75 & 21.75 & 6.39 & 1.02 \\
        \rowcolor{black!10}
        \cellcolor{white} 
        \textbf{\methodname} (Student) & C+R & 36.52(+0.73) & 11.67(+1.11) & 2.27(+0.52) & 23.66(+1.91) & 7.27(0.88) & 1.22(+0.2) \\
        \bottomrule
      \end{tabular}
    }
    \caption{
    Performance breakdown by range evaluated on the nuScenes \emph{val} split.
    }
    \label{tab:range}
\end{table*}
\begin{table*}[!htp]
\captionsetup{font={small}}
    \centering
    \resizebox{\linewidth}{!}{
     \begin{tabular}{c|c|cccc|cccc}
        \toprule
        & & \multicolumn{4}{c|}{IoU$\uparrow$} & \multicolumn{4}{c}{mIoU$\uparrow$}         \\
        \multirow{-2}{*}{Method} & \multirow{-2}{*}{Modality} & \includegraphics[height=0.15in]{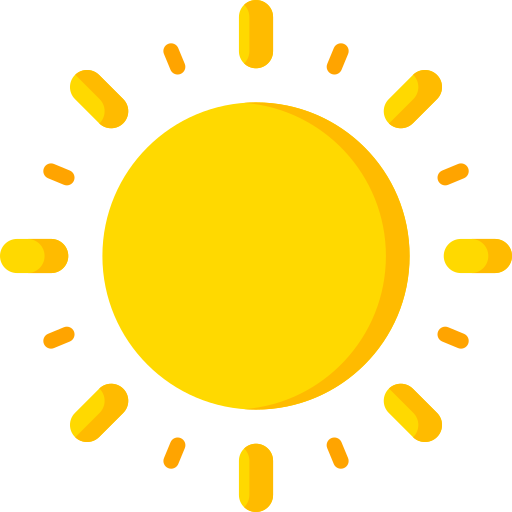}  & \includegraphics[height=0.15in]{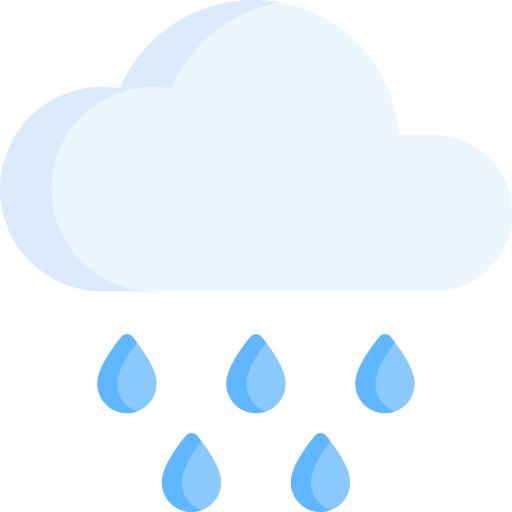}  & \includegraphics[height=0.15in]{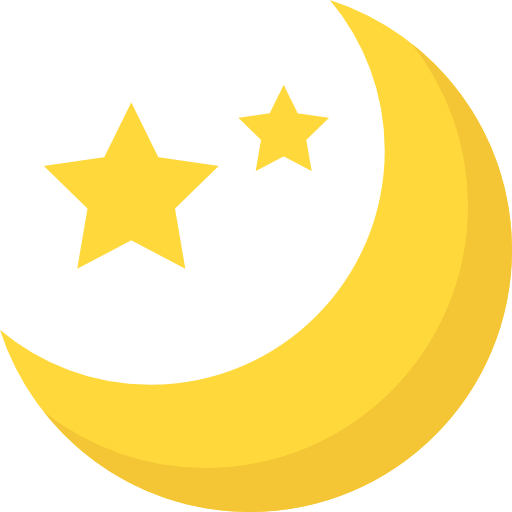} & \includegraphics[height=0.15in]{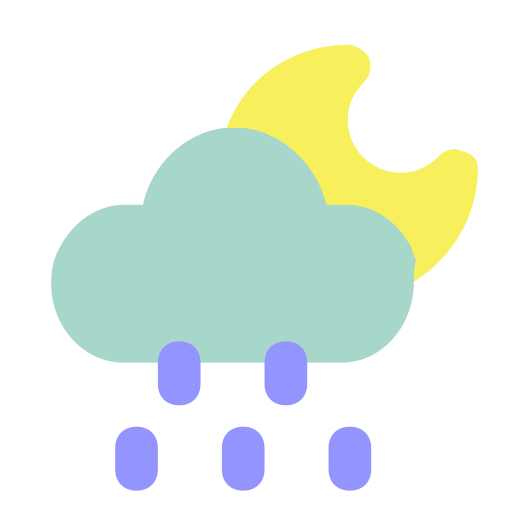} & \includegraphics[height=0.15in]{figure/sun.png}  & \includegraphics[height=0.15in]{figure/rain.png}  & \includegraphics[height=0.15in]{figure/moon.png} & \includegraphics[height=0.15in]{figure/RN.png} \\ 
        \midrule
        LC-Fusion (Teacher) & L+C  & 35.22 & 34.28  & 33.52 & 34.16 & 24.91 &  24.08  & 16.43 & 12.69\\ 
        SSC-RS & L  & 35.35 & 32.27  & 33.16 & 32.37 & 21.23 &  19.80  & 13.02 & 10.90\\
        C-CONet & C  & 23.20 & 19.90  & 9.80 & 9.60 & 14.10 &  12.70  & 4.80 & 3.64\\
        \midrule
        R-SSC-RS & R & 21.25 & 19.27  & 18.90 & 17.01 & 7.55 & 7.15  &  5.35 & 4.30 \\
        \rowcolor{black!10}
        \cellcolor{white} 
        \textbf{R-\methodname} (Student) & R & 21.86(+0.61) & 20.10(+0.83) & 19.30(+0.4) & 17.27(+0.26) & 8.59(+1.04) & 7.10(-0.05)  & 5.69(+0.34) & 4.74(+0.44) \\
        \midrule
        RC-Fusion & R+C & 25.88 & 24.49 & 20.47 & 20.10 & 15.85 & 15.30  & 9.39 &  7.10 \\
        \rowcolor{black!10}
        \cellcolor{white} 
        \textbf{RC-\methodname} (Student) & R+C  & {26.72(+0.84)} & 25.12(+0.63) & 21.23(+0.76) & 20.67(+0.57) & {17.48(+1.63)} & 16.69(+1.39)  & 10.10(+0.71) & 8.02(+0.92) \\
        \bottomrule
      \end{tabular}
    }
    \caption{
    Performance breakdown by weather and lighting evaluated on the nuScenes \emph{val} split. 
    }
    \label{tab:weather}
    \vspace{-10pt}
\end{table*}
\subsubsection{Weather Robustness from Radar}
\label{exp:Weather Robustness from Radar}
This study evaluates the performance of the radar-based methods in various weather conditions. Results detailed in Tab.~\ref{tab:weather} reveal that models' performances fluctuate with changing weather scenarios (sunny daytime, rainy day, nighttime, and rainy night). 

First, as shown in Tab.~\ref{tab:weather}, the weather properties of the three sensor types reveal varying degrees of robustness. The mIoU for radar decreases by only 3.25 from a clear day to a rainy night, while for LiDAR and camera, it decreases by 10.33 and 10.46 points, respectively. This indicates that radar is the most resilient to adverse weather and lighting conditions. In particular, during clear daylight hours, the teacher model achieves 16.32 higher than R-LiCROcc and 7.43 higher than RC-LiCROcc in terms of mIoU scores. However, in rainy night conditions, this advantage narrows to 7.95 and 5.26, respectively, with dominant performance decreasing by 51.3\% and 29.2\%.  Additionally, it is evident that the rain in the nuScenes dataset is not particularly heavy, resulting in a less significant impact on the LiDAR point cloud than anticipated. Examining radar performance under a broader range of weather conditions is a focus of our future work.

Under sunny daytime conditions, the distillation effect yields the highest performance. The R-LiCROcc model demonstrates a 2.8\% improvement in IoU and a 13.8\% improvement in mIoU compared to the student model. Similarly, the RC-LiCROcc model achieves a 3.2\% increase in IoU and a 10.3\% increase in mIoU. This enhancement is attributed to the optimal performance of the teacher model under sunny conditions. Conversely, during rainy days and nights, the visibility of both the LiDAR and camera is compromised, leading to less pronounced enhancements for the student model. In fact, the performance of the R-LiCROcc model is slightly diminished in rainy weather.

\section{Conclusions and Future Work}
\label{sec:conclustion}
In this paper, we investigate the utilization of radar in SSC task. We initially developed a fusion network that integrates point clouds and images, complemented by three distillation modules. By leveraging the strengths of radar while augmenting its performance on the SSC task, our approach achieves superior results across diverse settings. Moving forward, we plan to examine the SSC performance of additional radar types to substantiate our findings.
\section{Acknowledgement}
This work is partially supported by National Natural Science Foundation of China under grant U21A20484.


%

\ifCLASSOPTIONcaptionsoff
  \newpage
\fi



%

{
\AtNextBibliography{\scriptsize}
\printbibliography
}

%

\end{document}